\documentclass[a4paper]{./styles/svproc}
\usepackage{url}
\usepackage{graphicx}
\usepackage{wrapfig}
\usepackage{multirow}
\usepackage{hyperref}
\usepackage{amsmath}  
\usepackage{amssymb}  
\DeclareMathOperator*{\argmax}{arg\,max}

\usepackage[linesnumbered]{algorithm2e}
\SetKwComment{Comment}{// }{}
\RestyleAlgo{ruled}

\begin{document}
\mainmatter              
\title{Improving User Experience in Preference-Based Optimization of Reward Functions \\for Assistive Robots}
\titlerunning{Improving User Experience in Preference-Based Optimization}  
%
\author{Nathaniel Dennler \and Zhonghao Shi \and
Stefanos Nikolaidis \and Maja Matari\'c}
\authorrunning{Dennler, Shi, Nikolaidis, and Matari\'c} 
%
\tocauthor{Nathaniel Dennler, Zhonghao Shi, Stefanos Nikolaidis, Maja Matari\'c}
\institute{University of Southern California, Los Angeles CA 90007, USA,\\
\email{\{dennler, zhonghas, nikolaid, mataric\}@usc.edu}
}

\maketitle              

\begin{abstract}
Assistive robots interact with humans and must adapt to different users' preferences to be effective. An easy and effective technique to learn non-expert users' preferences is through rankings of robot behaviors, for example, robot movement trajectories or gestures. Existing techniques focus on generating trajectories for users to rank that maximize the outcome of the preference learning process. However, the generated trajectories do not appear to reflect the user's preference over repeated interactions. In this work, we design an algorithm to generate trajectories for users to rank that we call Covariance Matrix Adaptation Evolution Strategies with Information Gain (CMA-ES-IG). CMA-ES-IG prioritizes the user's experience of the preference learning process. We show that users find our algorithm more intuitive and easier to use than previous approaches across both physical and social robot tasks. This project's code is hosted at \href{https://github.com/interaction-lab/CMA-ES-IG}{github.com/interaction-lab/CMA-ES-IG}
\keywords{human-robot interaction, preference learning}
\end{abstract}
\section{Introduction}

Numerous technical advances in robotics have allowed robots to perform increasingly complex physical and social tasks. As robots move from laboratories to in-home settings, they will be confronted with users and contexts that were not previously seen or tested by the robot's developers. In order to be useful in these new contexts, robots must adapt their behaviors, e.g., movement trajectories, affective gestures, and voice, to align with preferences and expectations of specific users \cite{dennler2023design,rossi2017user}. One user may prefer that a robot hands them an item as quickly as possible, whereas another user may want the robot to stay as far away from a priceless family heirloom as possible. Users of these systems will likely not have experience directly programming robots, and thus robots must be teachable in more intuitive ways.

Previous work has identified that ranking robot trajectories can be a simple and effective method for non-expert users to teach robots their preferences for how the robot should move \cite{brown2020better,keselman2023optimizing}. Two main approaches use ranking information to learn preferences: (1) using rankings to learn an explicit model of a user's reward function \cite{brown2020better}, or (2) using rankings to implicitly infer a user's reward function through black-box optimization techniques \cite{keselman2023optimizing,lu2022preference}. These approaches have typically been evaluated in isolation and focused on the end-behavior that a robot has learned after the preference learning process has been completed \cite{biyik2019asking,sadigh2017active}. However, the process of teaching robots is a major factor that drives the perception and adoption of robotic systems~\cite{adamson2021we}.

\begin{figure}[t]
    \centering
    \includegraphics[width=\linewidth]{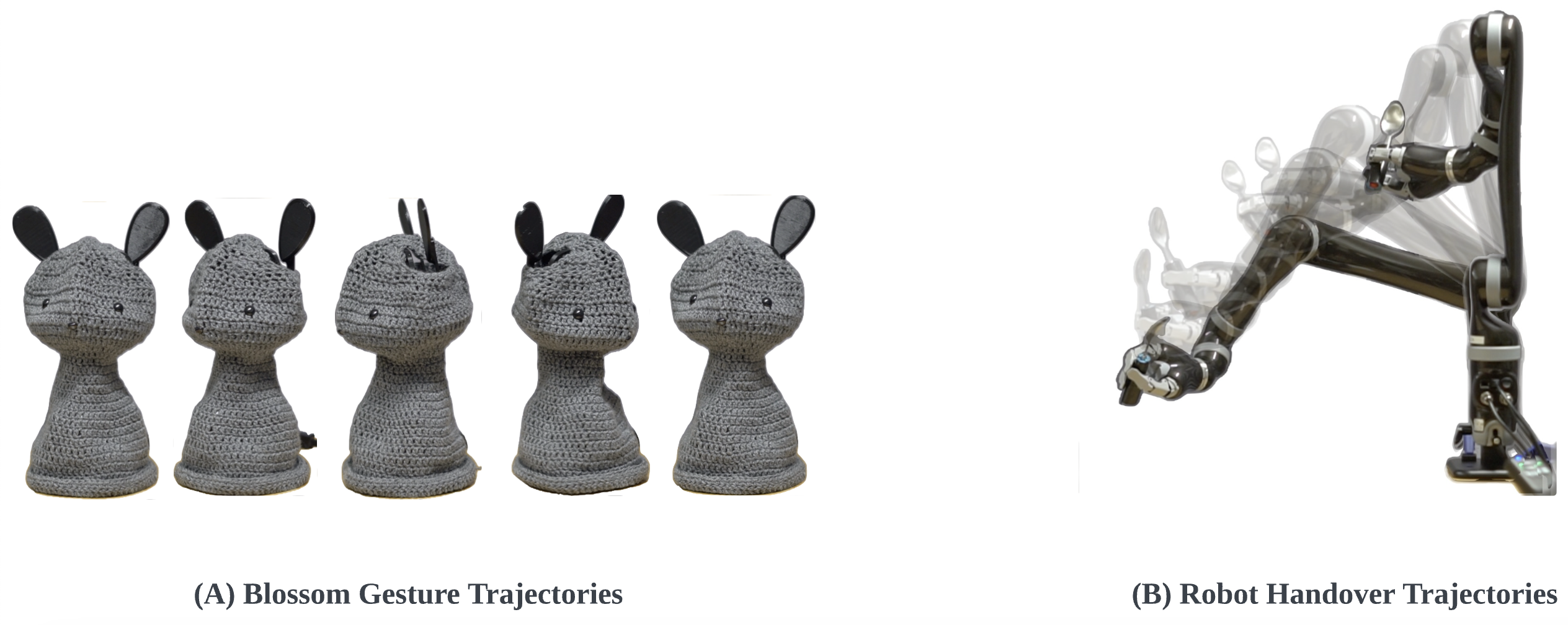}
    \caption{The two domains that users taught robots their preferences for the robot's behaviors. In the physical domain, users ranked a JACO arm's movement trajectories to hand them a marker, a cup, and a spoon. In the social domain, users ranked a Blossom robot's gestures to portray happiness, sadness, and anger. }
    \label{fig:tasks}
\end{figure}

In this work, we propose an algorithm that combines the explicit and implicit approaches for learning user preferences, called Covariance Matrix Adaptation Evolution Strategy with Information Gain (CMA-ES-IG). We show through simulation that CMA-ES-IG generates candidate robot trajectories that better reflect user preferences compared to state-of-the-art approaches. We then evaluate these algorithms through real-world experiments with users performing both physical and social robot tasks, as shown in Fig. \ref{fig:tasks}. Physically, users specified their preferences for how a JACO2 robot arm hands them various items. Socially, users specified their preferences for how a Blossom robot \cite{suguitan2019blossom,o2024design,shi2024build} gestures to communicate different emotional states. We show that our algorithm is able to effectively learn user preferences while also increasing the quality of the robot's trajectories over time. Overall, we highlight the importance of user experience in algorithmic design to create interactions that effectively learn user preferences.
 



\section{Related Work}

\subsubsection{Preferences in Physically Assistive Robots.}
Physically assistive robots help users with tasks such as collaborative assemblies \cite{nemlekar2023transfer,gombolay2018fast}, physical rehabilitation \cite{zhang2017human}, and activities of daily living \cite{bhattacharjee2020more,dennler2021design}. Though these goals appear to be objective and universal, differences in user preferences for how such goals are accomplished affect efficiency in human-robot collaboration \cite{zhang2017human,nikolaidis2015efficient}. For example, prior works have shown that automatic adaptation to user preferences by clustering users supports task completion \cite{nemlekar2023transfer,nikolaidis2015efficient}.

However, other works have found that users want to have some control over how a robot interacts with them. In particular, users can perceive fully autonomous physically assistive systems as a reduction of their own agency \cite{bhattacharjee2020more}. In order to restore user agency when interacting with these systems, researchers have employed personalization procedures in systems that assist users with eating \cite{canal2016personalization}, reaching for items \cite{jeon2020shared}, and handing items to the user \cite{perovic2023adaptive}, thereby increasing user perceptions of the system and the system's ability to perform tasks. In our study, we expand on robot handovers in the physical domain.

\subsubsection{Preferences in Socially Assistive Robots.} Socially assistive robots (SARs) are designed to assist users through primarily social interaction, rather than physical means \cite{mataric2016socially}. SARs have been used to facilitate behavior change in a wide variety of user populations, including college students \cite{o2024design,kian2024can}, children with learning differences \cite{saleh2021robot,shi2023evaluating}, users with limited mobility \cite{dennler2021personalizing,dennler2023metric}, and the elderly \cite{ghafurian2021social,zhou2021designing}. Due to the unique and varied needs of these populations, system designers must develop robot behaviors that are specifically tailored to each specific user population~\cite{clabaugh2019escaping}. 

Pragmatic constraints do not allow robot system developers to carefully design systems for each new user. Therefore, to make robots useful to diverse populations, users must be able to customize the system to align with their preferences \cite{gasteiger2023factors}. Previous works have found several benefits of adapting robot behaviors to the user \cite{rossi2017user}, though these works often require researchers to modify the robot \cite{martins2019user,tapus2008user}. Allowing the user to modify a robot's social behaviors such as voice \cite{shi2023evaluating}, personality \cite{tapus2008user}, challenge level of the assistive tasks \cite{clabaugh2019long}, and gesture \cite{moro2018learning} leads to increased acceptance of the system. 

\subsubsection{Learning User Preferences.}
There are many ways to learn preferences from users in order to support users in accomplishing tasks \cite{fitzgerald2022inquire}. Robots can learn a user's internal reward function through inverse reinforcement learning a function that maps a low-dimensional representation to a scalar value for each robot behavior. Various techniques can be used to learn this mapping from trajectory to reward, including demonstrations \cite{abbeel2004apprenticeship}, physical corrections \cite{bajcsy2017learning}, language \cite{sharma2022correcting}, trajectory rankings \cite{brown2020better}, and trajectory comparisons \cite{sadigh2017active}. Users may need varying levels of expertise with using robot systems to effectively use those techniques \cite{fitzgerald2022inquire}. Our work focuses on using {\it trajectory rankings}, a technique that is accessible to users of all levels of expertise with using robot systems.

There are two popular approaches for learning preferences when using rankings to learn user preferences for robot trajectories. One approach is to explicitly model the user's reward function by estimating the probability distribution over parameters of a reward function \cite{sadigh2017active,biyik2018batch}. This type of approach decomposes a trajectory ranking into a series of trajectory selections and uses probabilistic models of human choice to update the distribution over reward weights using Bayes' rule. The Bayesian approach considers the whole space of trajectories at each iteration of querying the user. By explicitly modeling users' reward functions, robots have been able to maneuver cars in simulation \cite{sadigh2017active} and assist with assembly tasks \cite{nemlekar2023transfer}. 

The other approach of interest is to implicitly model the user's reward function using rankings as inputs to black-box optimization algorithms. This type of approach directly finds the robot trajectories that the user will rank highly and dynamically updates the space of trajectories that are sampled to be ranked by the user. CMA-ES \cite{hansen2003reducing} is one technique that demonstrates efficiency and tolerance to noise, and has been applied to learning human user preference in robotics domains. CMA-ES has been applied to optimize and identify personalize control laws for exoskeleton assistance that minimizes human energy cost during walking~\cite{zhang2017human}. Recent work also applied CMA-ES to enable user preference learning through pairwise selection to enable user preference and goals learning in social robot navigation~\cite{keselman2023optimizing}.

Both types of approaches show promise in eliciting user preferences, but many prior works have focused on using these methods in isolation. Additionally, these approaches have focused on the outcome of the robot behavior, rather than on how users perceive the robot to be considering their input. A comprehensive work by Habiban et al. \cite{Habibian2022Heres} showed that incorporating users' perception is crucial for robots learning preferences over interpretable hand-crafted features.  In this work, we combine explicit and implicit user reward models to learn preferences for features we learned from data, and we evaluate users' perceptions of the teaching process.


\section{Learning User Preferences Through Comparisons}

\subsubsection{Preliminaries.} We describe robot behaviors as output {\it trajectories} from a dynamical system. A trajectory $\xi \in \Xi$ is defined as a sequence of states, $s \in S$ and actions $a \in A$ that follow the system dynamics, i.e, $\xi = (s_0,a_0,s_1,a_1,...,s_T,a_T) $ for a finite horizon of $T$ time steps. These states are abstractly defined and can be anything from joint angles to end-effector positions to images, and actions simply convert one state to another state. Following common practices in inverse reinforcement learning \cite{abbeel2004apprenticeship}, we assume that there exists a feature function $\phi: X \mapsto \mathbb{R}^d$ that represents aspects of the state that the user may have preferences over. A trajectory can then be represented as a low-dimensional vector in $\mathbb{R}^d$ via $\Phi(\xi) = \sum_{i=0}^T \phi(s_i)$. 

A user's preference for robot behaviors is a function of these trajectory features, $R(\xi) = f(\Phi(\xi))$. In this work, we make the assumption that a user's preference is a linear combination over features $R(\xi) = \omega^T \cdot \Phi(\xi)$, where $\omega \in \mathbb{R}^d$, as in several previous works \cite{sadigh2017active,biyik2019asking}. When a user is asked to rank trajectories, they are presented with a set of $N$ trajectories referred to as a \textit{trajectory query}, $Q = \{\xi_0,\xi_1,...,\xi_N\}$. A user then ranks these trajectories according to their internal reward function $\mathcal{R} = (\xi'_0, \xi_1', ..., \xi'_N)$ such that $\xi'_0 \prec \xi_1' \prec ... \prec \xi'_N$.

\subsubsection{Bayesian Optimization of Preferences.}
Approaches that explicitly model the user's reward function maintain a probability distribution over $\omega$ and update this distribution based on the user's ranking. A common framing is to view a ranking as an iterative process of selecting the best trajectories from the query without replacement \cite{myers2022learning}, and use the widely-adopted Bradley-Terry model of rational choice \cite{bradley1952rank} to estimate the probability that a user will select a given trajectory from a set of trajectories at each iteration, subject to a rationality parameter $\beta$:

\begin{equation}
p(\xi \mid Q) = \frac{e^{\beta \cdot \Phi(\xi)}}{\sum\limits_{\xi' \in Q}e^{\beta \cdot \Phi(\xi')}}
\end{equation}

Using this model of user preferences and assuming that these selections are conditionally independent, the distribution over $\omega$ can be calculated using Bayes' rule:

\begin{equation} \label{eq:weight_update}
p(\omega \mid \mathcal{R}) \propto p(\omega) \prod_{i = 0}^N p(\xi_i \mid Q_i)
\end{equation}

where $\xi_i$ represents the trajectory selected from the set at iteration $i$ and $Q_i$ represents the set of trajectories left at iteration $i$. Equivalently, $Q_{i} = Q_{i+1} \cup \xi_i$. 

The state of the art technique for generating the set of trajectories Q that a user ranks is to maximize an information gain objective as described by Bıyık et al. \cite{biyik2019asking}:

\begin{equation}\label{eq:infogain}
Q = \argmax_{Q = \{\xi_0, \xi_1,...,\xi_N\}} H(\omega \mid Q) - \mathbb{E}_{\xi \in Q} H(\omega \mid \xi, Q)
\end{equation}

where H is the Shannon Information Entropy. Maximizing this objective to form Q results in a set of trajectories that maximally update the distribution over $\omega$ when receiving the user's feedback. In addition, these trajectories are distinct from each other, enabling the user to easily differentiate among them. For linear reward functions, this implies that trajectories have large distances from each other in feature space.

\subsubsection{Covariance Matrix Adaptation Evolution Strategies (CMA-ES).}

Evolution strategies (ES) is a large body of algorithms that focus on solving continuous, black-box, mainly experimental optimization problems. ES algorithms sample a population of solutions for each generation, and move this sampled population toward solutions with more optimal fitness values generation by generation~\cite{back1997evolutionary}. Covariance Matrix Adaptation Evolution Strategies (CMA-ES) is proposed to reduce the number of generations needed to converge to the optimal solutions and improve the noise tolerance of the optimization process~\cite{hansen2003reducing}. Compared to other ES methods, CMA-ES has been evaluated as one of the most competitive derivative-free optimization algorithms for continuous spaces~\cite{hansen2010comparing}. More specifically, CMA-ES samples an underlying distribution of the population from a multivariate normal distribution, defined as $\mathcal{N}(m, C)$, where $m \in \mathbb{R}^d$ is the distribution mean and $C \in \mathbb{R}^{d \times d}$ is the symmetric and positive definite covariance matrix for the distribution~\cite{hansen2016cma}. At each step of CMA-ES, we sample trajectories from this distribution, and return the ranked values to the CMA-ES optimizer. The optimizer updates $m$ and $C$ using the CMA-ES update algorithm to move the distribution to higher-valued areas in the trajectory feature space.

In practice, CMA-ES samples narrower and higher performing regions of the trajectory feature space after each iteration. This results in an increasing average reward for the trajectories that the user is asked to rank, increasing the user's perception that the system is actually learning their preferences. However, because it samples a normal distribution, it often presents users with trajectories that are too similar to each other for a user to easily differentiate between them.

\subsubsection{Combining Information Gain and CMA-ES}

\begin{figure}[t]
    \centering
    \includegraphics[width=.7\linewidth]{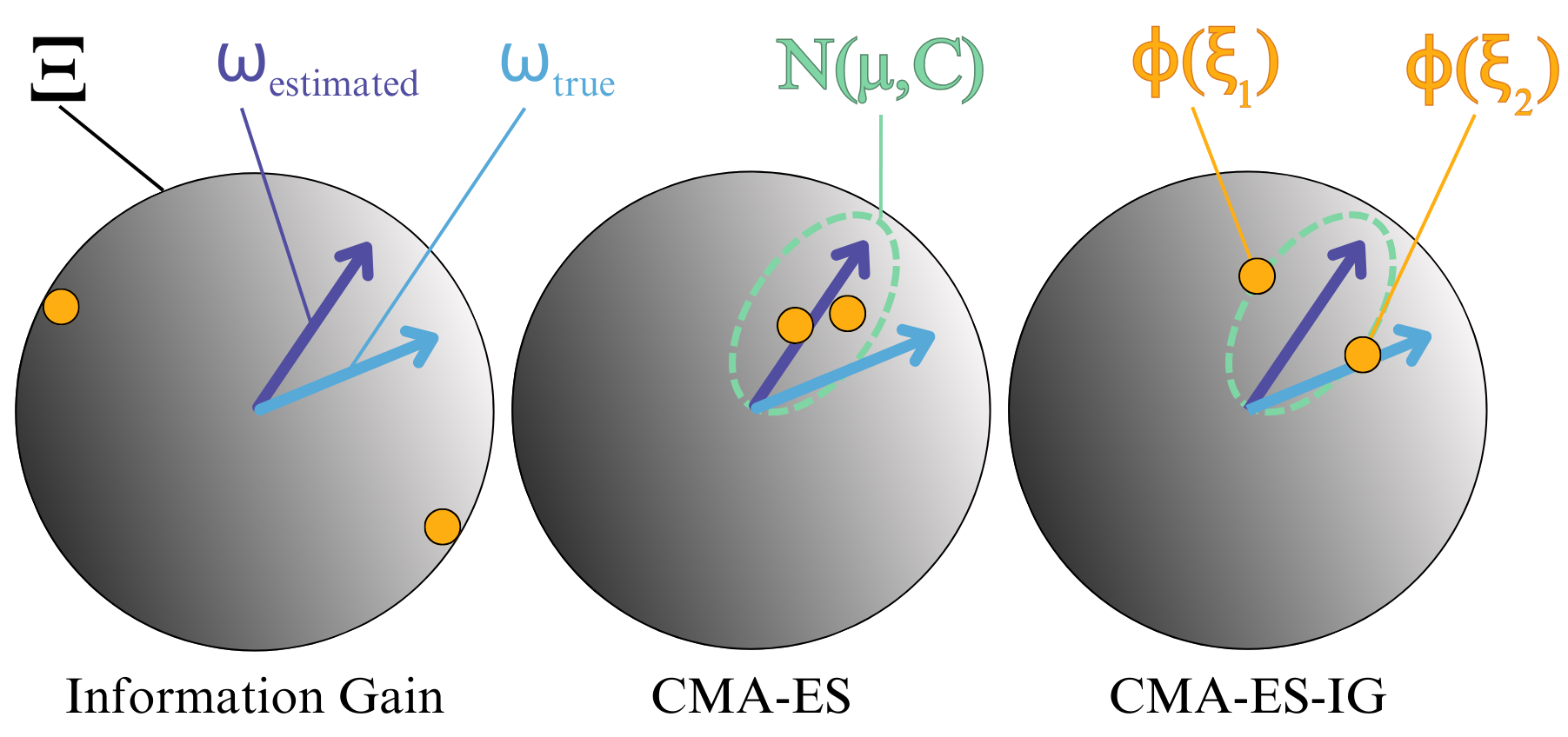}
    \caption{Example queries generated from an early step of each algorithm. The large circle represents the space of all trajectories with lighter areas representing higher reward, light blue arrows representing the user's true preference, dark blue arrows representing the current estimate of the user's preference, orange circles representing sampled trajectories to present to the user, and green dotted regions representing the sampling distribution from the current step of the CMA-ES optimizer. Information gain results in easy to differentiate queries, CMA-ES results in higher rewards on average, and CMA-ES-IG results in higher rewards that are easy to differentiate.}
    \label{fig:visual_intuition}
    \vspace{-1em}
\end{figure}

We propose a new algorithm, Covariance Matrix Adaptation Evolution Strategies with Information Gain (CMA-ES-IG), for efficiently learning user preferences. CMA-ES-IG leverages the benefits of both of the previous strategies: it uses the information gain objective to generate sets of trajectories that are easy to rank, and it uses the adaptive sampling mechanism from CMA-ES to increase the average user reward of the proposed trajectories over time. We summarize CMA-ES-IG in Algorithm \ref{alg:cmaesig}, and provide a visual intuition of these three algorithms in Fig. \ref{fig:visual_intuition}.

First, we initialized CMA-ES-IG identically to CMA-ES and our belief over user preferences to a uniform distribution. We then sample $D$ trajectory features from the CMA-ES-IG mean and covariance to create a set of trajectory features, $\mathcal{D}$. Next, we find $|Q|$ samples from $\mathcal{D}$ that maximize the expected information gain, as described in Equation \ref{eq:infogain}. While finding the exact solution to this optimization problem is exponential in $|Q|$, an efficient approximation is to find $|Q|$ medoids in the set of samples \cite{biyik2018batch}. We adopt this approximation to allow CMA-ES-IG to be computationally tractable.

\begin{algorithm}[t]
\caption{CMA-ES-IG}\label{alg:cmaesig}
\textbf{Given} a dataset of robot trajectories $\mathcal{D}$, a function that generates trajectory features $\Phi$, a number of items to ask the user $|Q|$, and a prior belief over user preferences $b_0$\;

\textbf{Initialize} the CMA-ES optimizer with $\mu,C$\; 

\While{user not done}{
$S \gets$ D samples from $\mathcal{N}(\mu, C)$\;
$\Omega \gets$ D samples from $b_t$\;
$Q \gets \arg\max_{ \{q_1,q_2,...,q_{|Q|} \mid q \in S \} } \sum_{\omega \in \Omega} H(q \mid \omega, Q) - \mathbb{E}_\omega (q \mid \omega, Q)$\;
$R \gets $ User's ranked responses\; 
$b_{t+1} \propto b_t \prod_{i = 0}^N p(\xi_i \mid Q_i) $\;
Update $\mu,C$ according to CMA-ES update \cite{hansen2003reducing}\;
}

\end{algorithm}

\subsection{Validation through Simulated Preferences}
To validate our algorithm before presenting it to users, we performed an algorithmic analysis by simulating user preferences as in previous work \cite{sadigh2017active,biyik2019asking}. We used the parameter estimation task as described by Fitzgerald et al. \cite{fitzgerald2022inquire}, which samples a ground truth weight vector, $\omega^*$ from a d-dimensional space. To ensure that each $\omega^*$ is comparable, we projected each to the unit ball \cite{sadigh2017active}. We then generated queries of four trajectories using our three algorithms (Infogain, CMA-ES, and CMA-ES-IG) for a simulated user to rank. We used the distribution described by Bradley-Terry preference model to perform these rankings, given the ground truth vector of the simulated user. We updated the distribution over the weight vector using Equation \ref{eq:weight_update}. We simulated 100 users performing 30 rankings for each of the three algorithms.

\begin{figure}[ht]
    \centering
    \includegraphics[width=\linewidth]{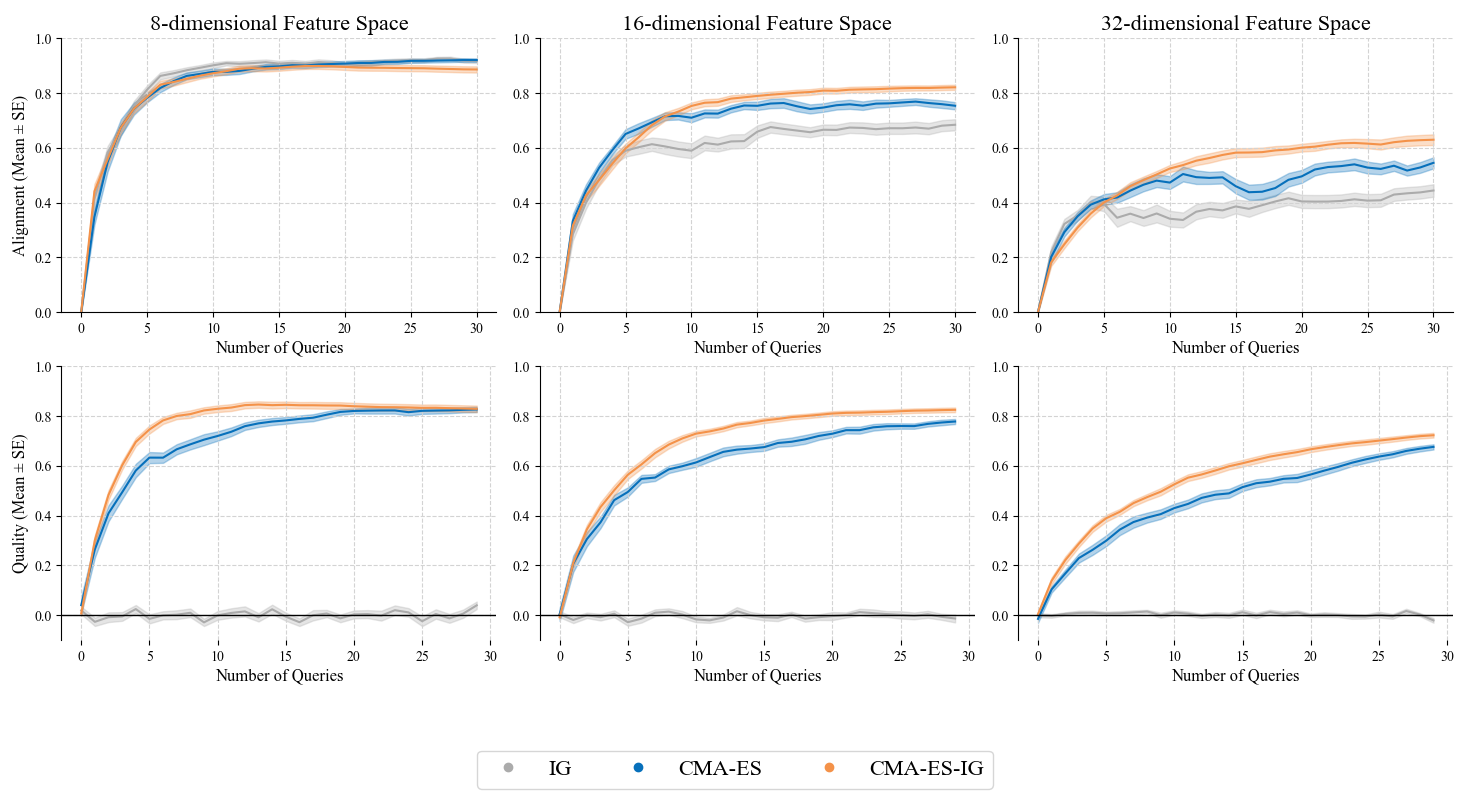}
    \caption{Comparison of simulation results for learning user preferences. Shaded regions indicate standard error. We found that all methods were able to learn user preferences across varying dimensions. The quality of the trajectories in the query increases only for CMA-ES and CMA-ES-IG, with CMA-ES-IG performing significantly better.}
    \label{fig:simulation_results}
\end{figure}

We are interested in two evaluation metrics: \textit{alignment}, which measures how well the estimated preference matches the true preference, and \textit{quality}, which measures the overall reward of the trajectories presented to the user. We define alignment as the cosine similarity between the estimated $\omega^{est}$ and the ground truth $\omega^*$, as in previous works \cite{fitzgerald2022inquire,sadigh2017active}. We define quality as the average reward of the trajectories in the query $\frac{1}{\mid Q \mid}\sum_{i=0}^{\mid Q \mid} \omega^* \cdot \Phi(\xi_i)$. We measure these values after each simulated query to generate curves that show how each metric increases with repeated querying, as shown in Fig.~\ref{fig:simulation_results}. To compare between curves, we used the area under the curve (AUC) metric, which provides values between -1 and 1, with 1 being the best. We show results for \textit{alignment} and \textit{quality} across features spaces of $d \in \{8,16,32\}$ in Table \ref{tab:quantitative_results}; For completeness, we also report regret as a secondary metric for \textit{alignment} in Table \ref{tab:quantitative_results}. Where $regret = \omega^*\cdot\phi(\xi^*) - \omega^*\cdot\phi(\xi')$; $\xi^*$ denotes the trajectory with the highest reward in $\Xi$ under $\omega^*$, and $\xi'$ denotes trajectory with the highest reward in $\Xi$ under $\omega^{est}$.

\begin{table}[]
\centering
\caption{Quantitative Results. We report the area under the curve (AUC) for alignment of learned reward and quality of query across d-dimensional feature spaces.}
\label{tab:quantitative_results}
\begin{tabular}{r|ccc|ccc|ccc}
\hline
       & \multicolumn{3}{c}{Alignment ($\uparrow$)} & \multicolumn{3}{c|}{Regret ($\downarrow$)}  & \multicolumn{3}{c}{Quality ($\uparrow$)} \\
          & $d=8$ & $d=16$ & $d=32$ & $d=8$ & $d=16$ & $d=32$ & $d=8$ & $d=16$ & $d=32$ \\ \hline
IG     & \textbf{.848} & .606 & .374 & \textbf{.331}  & 1.243 & 2.115 & -.001  & -.003  & .003  \\
CMA-ES & .834 & .691 & .488 & .479 & .995 & 1.876 & .688 & .601 & .450 \\
CMA-ES-IG & .828 & \textbf{.717} & \textbf{.517} & .393 & \textbf{.759} & \textbf{1.453} & \textbf{.746} & \textbf{.673} & \textbf{.527} \\ \hline
\end{tabular}
\end{table}

We compared the three algorithms: Information Gain (IG), CMA-ES, and CMA-ES-IG. For the alignment metric, we saw that all three methods performed similarly after thirty queries, except for high-dimensional feature spaces. However, for the quality metric, we found that CMA-ES-IG outperformed CMA-ES and both evolutionary strategies vastly outperformed the IG algorithm. This indicates that all algorithms can learn user preferences, but only CMA-ES-IG and CMA-ES are effective at presenting robot behaviors that appeal to the user through interaction.

\section{Experimental Evaluation}

To evaluate user perception of different methods for optimizing preferences, we conducted a within-subjects user study to compare participant perceptions among the different methods for learning preferences. In the physical domain, participants specified preferences for how a JACO robot arm hands them a marker, cup, or spoon. In the social domain, participants specified preferences for how a Blossom robot performed affective gestures to communicate happiness, sadness, and anger.

\subsection{Experimental Setup}

\begin{figure}
    \centering
    \includegraphics[width=\linewidth]{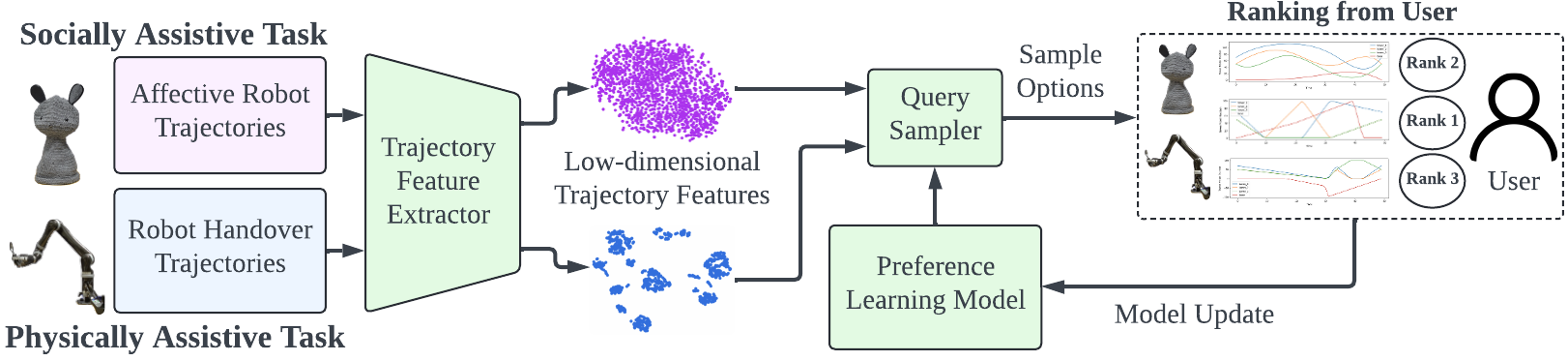}
    \caption{The framework for learning user preferences. We learned nonlinear features for sets of robot trajectories. The query sampler produced sets of trajectories for the user to rank and those rankings were used to update the estimate of the user's preferences.}
    \label{fig:framework}
\end{figure}

We developed a framework, shown in Fig.~\ref{fig:framework}, to evaluate the different techniques for learning participant preferences. It consists of three learned components: trajectory representations, query sampling, and preference learning models.

To generate trajectory representations, we used a dataset of 1000 robot handover trajectories and 1500 Blossom gesture trajectories. Trajectories consisted of a sequence 50 joint states that were equally spaced in time. The Blossom robot has joint states in $\mathbb{R}^4$, and trajectories were played by sending servo commands to each of the four servos that control the Blossom at a rate of 10Hz. The JACO arm has joint states in $\mathbb{R}^6$ so trajectories were played by fitting a b-spline to the those joint states and sending the goal joint states to an impedance controller at a rate of approximately 50Hz. The JACO arm trajectories were scaled to 9 seconds in duration. We generated trajectories for Blossom and the JACO arm by sampling from demonstrations, however other techniques can be used to create the datasets, such as quality diversity, reinforcement learning, or planning approaches. 

\begin{figure}[t]
    \centering
    \includegraphics[width=.7\linewidth]{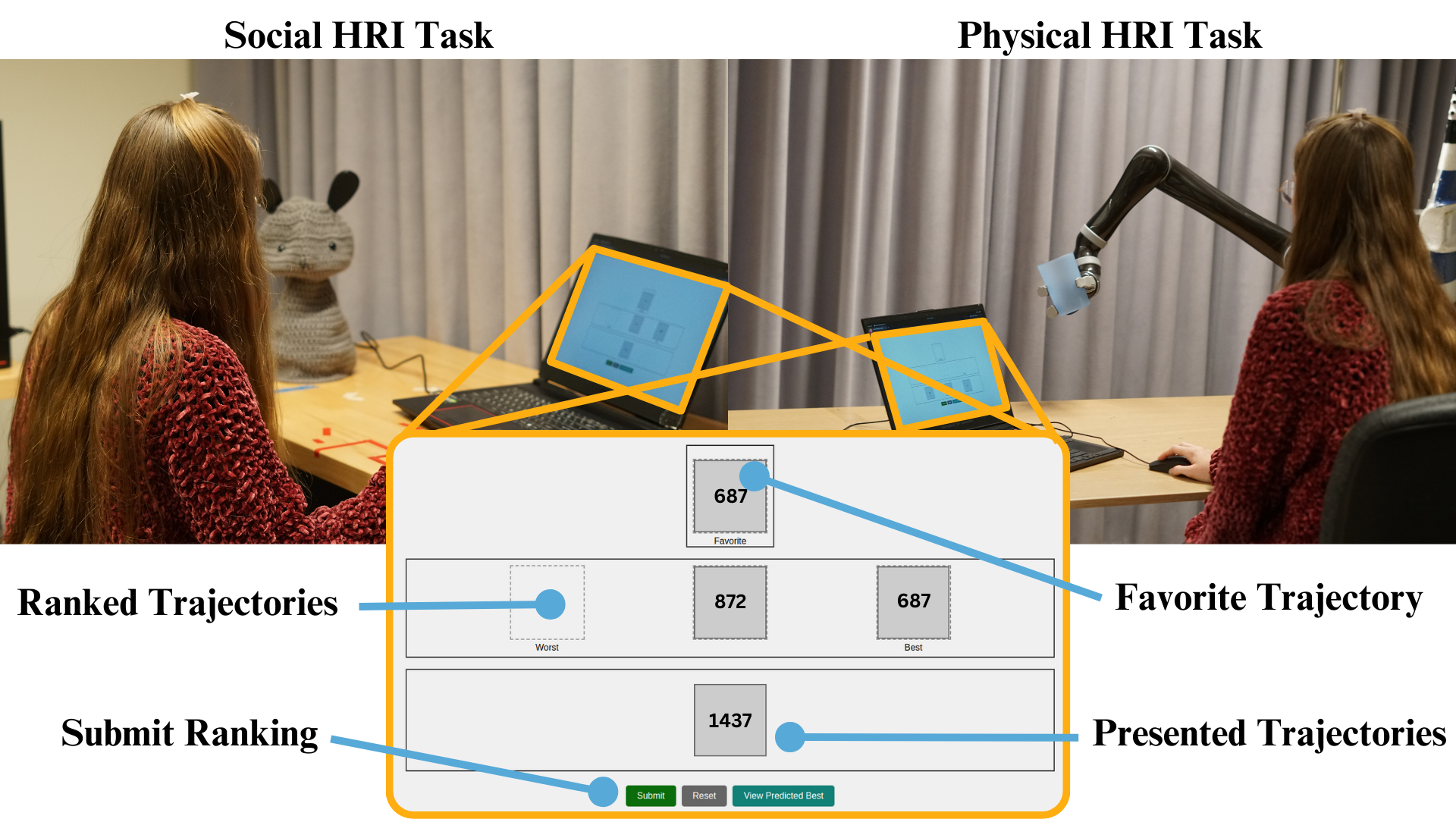}
    \caption{User study setup. Users interacted with the robots through the ranking interface to specify their preferences for how the Blossom robot used gestures to signal different affective states and how the JACO robot arm handed them different items.}
    \label{fig:study_setup}
\end{figure}

From the generated trajectories, we used an autoencoder (AE) to generate nonlinear features for the trajectories for the social and physical domains in our study, which is shown to be effective in prior work \cite{brown2020safe}. The AEs consisted of three convolutional layers and two fully connected layers with leaky ReLU activations after each layer. Hyperparameters were tuned to minimize the reconstruction loss over all trajectories in the dataset. The feature space of Blossom's trajectories was six-dimensional and the features space of the JACO arm trajectories was four-dimensional.

The query sampling component generated the set of trajectories to show to the participant. At that point we performed our experimental intervention. Participants were randomly presented with either the IG, CMA-ES, or CMA-ES-IG methods of generating sets of three robot behaviors to rank. We then identified the closest trajectories in the dataset to the sampled trajectory features to present to the participant. All methods took less than one second to calculate the set of presented behaviors. The participant ranked these three trajectories using the interface shown in Fig.~\ref{fig:study_setup}.

The interface presented trajectories to the participant, and the participant clicked on boxes to view each trajectory suggested by the algorithms. The participant dragged the boxes to the ranking area, placing the lowest-ranked trajectory in the leftmost box and the highest ranked trajectory in the rightmost box. If a participant found a trajectory particularly aligned with their preferences, they could place it in the ``Favorite" trajectory box. The trajectory in this box was saved for the participant to refer to for the entire interaction and could be replaced if they found a better trajectory. This box was informed by a pilot study where participants requested a way to refer to trajectories they saw previously. Once the participant ranked the three options, they selected the submit button to update the model of their preferences. At any time, they could also press the ``View Predicted Best'' button to view the trajectory that maximized their estimated reward.

Before each task, the preference learning algorithm was initialized to a uniform distribution. The distribution was updated according to Equation \ref{eq:weight_update} after each ranking. We used the Bradley-Terry model to represent the way the user ranked these trajectories, however other models can be used depending on the inputs available to the participant. For example previous work has proposed choice models that allows users to rank options as ``approximately equal" \cite{biyik2019asking}. After the participant provides their feedback and the model is updated, the query sampling process begins again with a new estimate of the participant's preferences, and the cycle repeats until the participant is satisfied.

\begin{table}[ht]
\caption{The Likert scale items that participants answered for our two metrics: perceived ease of use and perceived behavioral adaptation.}
\vspace{.2cm}
\label{tab:constructs}
\resizebox{\textwidth}{!}{%
\begin{tabular}{p{0.7\textwidth} @{\hspace{0.3cm}} | p{0.25\textwidth}}
\hline
\vspace{.05cm}
1. Teaching the robot is clear and understandable. & \multirow{4}{=}{\parbox{0.2\textwidth}{\centering \textbf{Perceived Ease of Use \cite{venkatesh2000theoretical}}}} \\ 
2. Teaching the robot does not require a lot of mental effort. & \\ 
3. I find the robot easy to teach. & \\ 
4. I find it easy to get the robot to do what I want it to do. \vspace{.1cm}& \\ 
\hline
\vspace{.05cm}
1. The robot has developed its skills over time because of my interaction with it. & \\
2. The robot's behavior has changed over time because of my interaction with it. & \multirow{4}{0.2\textwidth}{\centering \textbf{Perceived Behavioral Adaptation \cite{lee2005can}}} \\ 
3. The robot has become more competent over time because of my interaction with it. & \\ 
4. The robot's intelligence has developed over time because of my interaction with it. \vspace{.1cm} & \\ 
\hline
\end{tabular}%
}
\vspace{-.2cm}
\end{table}

\subsection{User Study Details}

We performed a within-subjects user study to identify the differences in user experience between the proposed algorithms across two domains. In particular, we were interested in two key factors that determine the actual use of systems: ease of use \cite{venkatesh2000theoretical} (EOU), and perceived behavioral adaptation \cite{lee2005can} (BA). Perceived ease of use measures how easily participants are able to get the robot to do what they want, and behavioral adaptation measures how much the users perceive the robot as changing in response to their inputs.
Specific Likert scale items in our study are listed in Table~\ref{tab:constructs}. Users rated these metrics on a 9-point Likert scale with 0 corresponding to strongly disagree and 8 corresponding to strongly agree. We average across the questions for each of these factors to calculate our evaluation metric.

Our study procedure was approved by the University of Southern California IRB under \#UP-24-00600 and proceeded as follows: first, the participant was greeted by the experimenter and randomly assigned to specify their preferences for either the physical or social robot interaction. Next, the participant specified their preferences using a randomized and counter-balanced algorithm for the first task for their assigned domain--a marker handover in the physical domain, or a happy gesture in the social domain. After five minutes, the participant rated the algorithm's EOU and BA. Next, the participant was presented with the next randomized algorithm and completed the second task for their assigned domain--a cup handover in the physical domain, or sad gesture for the social domain. Participants then rated the EOU and BA of the algorithm. The participant then interacted with the final randomized algorithm in their assigned domain--a spoon handover in the physical domain, or an angry gestures for the social domain. The user rated the EOU and BA of the final algorithm. The user then ranked the three algorithms against each other to specify their overall preferences. The participant then completed the same process for the other domain.

All participants were compensated with a 10 USD Amazon giftcard sent to their email. We recruited 14 participants; they were aged 19-32 (Median = 24, SD = 4.5) and comprised 6 women, 5 men, and 3 nonbinary individuals. There were 7 Asian, 1 Black, 3 Hispanic, and 5 White participants (some participants were more than one ethnicity). 

\subsection{User Study Results}

\begin{wrapfigure}{r}{0.5\textwidth}
\vspace{-1.5cm}
  \centering
  \includegraphics[width=0.48\textwidth]{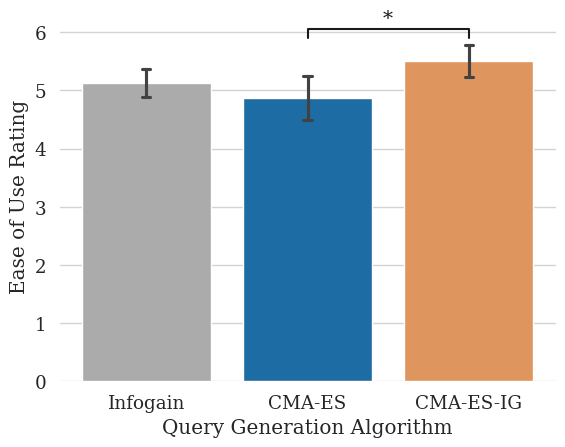}
  \caption{Ease of Use results. CMA-ES-IG was rated significantly easier to use than CMA-ES, and empirically easier than IG.}
  \vspace{-.8cm}
  \label{fig:EOU_results}
\end{wrapfigure}

\subsubsection{Ease of Use.}
We evaluated the average scores for EOU from a four-item Likert scale. We identified high internal consistency of the scale, with a Cronbach's alpha of $\alpha=.89$, indicating good internal consistency. To evaluate significance we used pairwise non-parametric repeated-measures tests. As shown in Fig.~\ref{fig:EOU_results}, we found that CMA-ES-IG received the highest EOU ratings ($M=5.50$), followed by Information Gain ($M=5.13$), and CMA-ES received the lowest rating for EOU ($M=4.87$). The difference between ratings for CMA-ES-IG and CMA-ES was significant ($W=5.5$, $p=.016$) with a medium effect size (hedge's $g=.558$). This indicates that including the information gain objective in CMA-ES-IG indeed makes it easier to use.

\begin{wrapfigure}{r}{0.5\textwidth}
  \centering
  \vspace{-1cm}
  \includegraphics[width=0.48\textwidth]{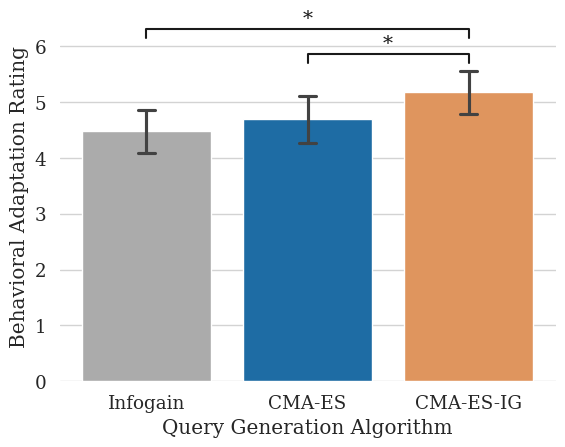}
  \caption{Behavioral Adaptation results. CMA-ES-IG was rated as changing the robot's behavior significantly more over time than both CMA-ES and IG.}
  \vspace{-0.8cm}
  \label{fig:BA_results}
\end{wrapfigure}

\subsubsection{Perceived Behavioral Adaptation.}
We evaluated the average scores for BA from a four-item Likert scale, and identified high internal consistency of the scale, with a Cronbach's alpha of $\alpha=.97$, indicating excellent internal consistency. We used non-parametric repeated-measures pairwise comparisons to assess significance. As shown in Fig.~\ref{fig:BA_results}, we found that CMA-ES-IG received the highest BA ratings ($M=5.18$), followed by CMA-ES ($M=4.69$), and Information Gain received the lowest rating for BA ($M=4.48$). CMA-ES-IG was rated as significantly higher than both CMA-ES ($W=15$, $p=.033$) with a small to medium effect (hedge's $g=.377$), and IG ($W=5.5$, $p=.009$) with a medium effect size (hedge's $g=.414$). This indicates that participants observed the largest behavioral change during the teaching process when using CMA-ES-IG to teach the robots.

\begin{wrapfigure}{r}{0.5\textwidth}
  \centering
  \vspace{-1cm}
  \includegraphics[width=0.48\textwidth]{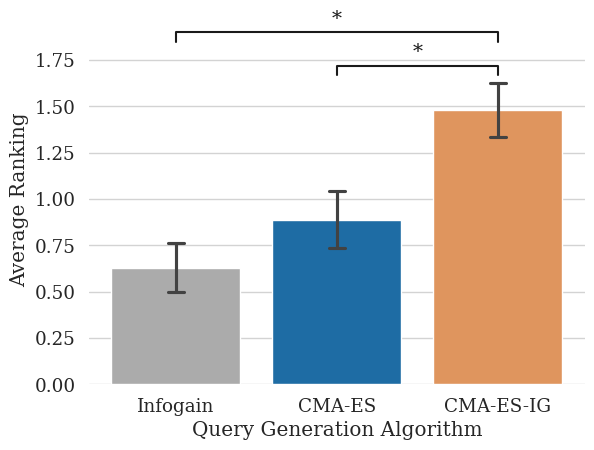}
  \caption{Algorithmic ranking results. CMA-ES-IG was consistently ranked as the most preferred algorithm for teaching robots preferences in our user study.}
  \label{fig:rank_results}
  \vspace{-.7cm}
\end{wrapfigure}

\subsubsection{Overall Ranking.}
Finally, we evaluated the ranking of each algorithm based on how the participants ranked the algorithms against one another. A ranking of 0 indicates that the participant believed that algorithm to be the worst overall, and a ranking of 2 indicated that the participant believed that algorithm to be the best overall. As shown in Fig.~\ref{fig:rank_results}, we found that users ranked CMA-ES-IG as the best algorithm on average ($M=1.48$), followed by CMA-ES ($M=.89$), and IG as the least preferred ($M=.63$). Using pairwise non-parametric repeated-measures tests, we found that CMA-ES-IG was consistently ranked significantly higher than CMA-ES ($W=3.0$, $p=.022$) with a large effect size (hedge's $g=1.26$). CMA-ES-IG was also ranked higher than IG ($W=0.0$, $p=.009$) with a large effect size (hedge's $g=1.75$). This indicates that, overall, participants preferred to use CMA-ES-IG to teach robots through trajectory rankings.

\vspace{-1em}

\subsection{Conclusion}

In this paper, we identified how the user experience of teaching robots preferences can be improved. We proposed a novel algorithm, CMA-ES-IG, that presents options for users to rank that become increasingly better over time and are easy for users to differentiate. We showed through simulation that CMA-ES-IG robustly improves the quality of these options across several parameters while maintaining the ability to efficiently learn user preferences. We showed experimentally through a user study that participants teaching physical and social robots prefered to use CMA-ES-IG, found CMA-ES-IG the easiest to use, and found CMA-ES-IG the most adaptive algorithm compared to the current state of the art approaches.

\textbf{Acknowledgements.}
This work was funded in part by an Amazon Research Award and by the National Science Foundation Convergence Accelerator grant (ITE-2236320). We thank Shihan Zhao and Amy O'Connell for their feedback and discussions on this work. 

%
%
\bibliographystyle{styles/bibtex/splncs03}
\bibliography{main}

\begin{thebibliography}{10}
\providecommand{\url}[1]{\texttt{#1}}
\providecommand{\urlprefix}{URL }

\bibitem{abbeel2004apprenticeship}
Abbeel, P., Ng, A.Y.: Apprenticeship learning via inverse reinforcement learning. In: Proceedings of the twenty-first international conference on Machine learning. p.~1 (2004)

\bibitem{adamson2021we}
Adamson, T., Ghose, D., Yasuda, S.C., Shepard, L.J.S., Lewkowicz, M.A., Duan, J., Scassellati, B.: Why we should build robots that both teach and learn. In: Proceedings of the 2021 ACM/IEEE International Conference on Human-Robot Interaction. pp. 187--196 (2021)

\bibitem{back1997evolutionary}
Back, T., Hammel, U., Schwefel, H.P.: Evolutionary computation: Comments on the history and current state. IEEE transactions on Evolutionary Computation  1(1),  3--17 (1997)

\bibitem{bajcsy2017learning}
Bajcsy, A., Losey, D.P., O’malley, M.K., Dragan, A.D.: Learning robot objectives from physical human interaction. In: Conference on robot learning. pp. 217--226. PMLR (2017)

\bibitem{bhattacharjee2020more}
Bhattacharjee, T., Gordon, E.K., Scalise, R., Cabrera, M.E., Caspi, A., Cakmak, M., Srinivasa, S.S.: Is more autonomy always better? exploring preferences of users with mobility impairments in robot-assisted feeding. In: Proceedings of the 2020 ACM/IEEE international conference on human-robot interaction. pp. 181--190 (2020)

\bibitem{biyik2019asking}
B{\i}y{\i}k, E., Palan, M., Landolfi, N.C., Losey, D.P., Sadigh, D.: Asking easy questions: A user-friendly approach to active reward learning. arXiv preprint arXiv:1910.04365  (2019)

\bibitem{biyik2018batch}
Biyik, E., Sadigh, D.: Batch active preference-based learning of reward functions. In: Conference on robot learning. pp. 519--528. PMLR (2018)

\bibitem{bradley1952rank}
Bradley, R.A., Terry, M.E.: Rank analysis of incomplete block designs: I. the method of paired comparisons. Biometrika  39(3/4),  324--345 (1952)

\bibitem{brown2020safe}
Brown, D., Coleman, R., Srinivasan, R., Niekum, S.: Safe imitation learning via fast bayesian reward inference from preferences. In: International Conference on Machine Learning. pp. 1165--1177. PMLR (2020)

\bibitem{brown2020better}
Brown, D.S., Goo, W., Niekum, S.: Better-than-demonstrator imitation learning via automatically-ranked demonstrations. In: Conference on robot learning. pp. 330--359. PMLR (2020)

\bibitem{canal2016personalization}
Canal, G., Aleny{\`a}, G., Torras, C.: Personalization framework for adaptive robotic feeding assistance. In: Social Robotics: 8th International Conference, ICSR 2016, Kansas City, MO, USA, November 1-3, 2016 Proceedings 8. pp. 22--31. Springer (2016)

\bibitem{clabaugh2019long}
Clabaugh, C., Mahajan, K., Jain, S., Pakkar, R., Becerra, D., Shi, Z., Deng, E., Lee, R., Ragusa, G., Matari{\'c}, M.: Long-term personalization of an in-home socially assistive robot for children with autism spectrum disorders. Frontiers in Robotics and AI  6,  110 (2019)

\bibitem{clabaugh2019escaping}
Clabaugh, C., Matari{\'c}, M.: Escaping oz: Autonomy in socially assistive robotics. Annual Review of Control, Robotics, and Autonomous Systems  2(1),  33--61 (2019)

\bibitem{dennler2023metric}
Dennler, N., Cain, A., De~Guzman, E., Chiu, C., Winstein, C.J., Nikolaidis, S., Matari{\'c}, M.J.: A metric for characterizing the arm nonuse workspace in poststroke individuals using a robot arm. Science Robotics  8(84),  eadf7723 (2023)

\bibitem{dennler2023design}
Dennler, N., Ruan, C., Hadiwijoyo, J., Chen, B., Nikolaidis, S., Matari{\'c}, M.: Design metaphors for understanding user expectations of socially interactive robot embodiments. ACM Transactions on Human-Robot Interaction  12(2),  1--41 (2023)

\bibitem{dennler2021design}
Dennler, N., Shin, E., Matari{\'c}, M., Nikolaidis, S.: Design and evaluation of a hair combing system using a general-purpose robotic arm. In: 2021 IEEE/RSJ International Conference on Intelligent Robots and Systems (IROS). pp. 3739--3746. IEEE (2021)

\bibitem{dennler2021personalizing}
Dennler, N., Yunis, C., Realmuto, J., Sanger, T., Nikolaidis, S., Matari{\'c}, M.: Personalizing user engagement dynamics in a non-verbal communication game for cerebral palsy. In: 2021 30th IEEE International Conference on Robot \& Human Interactive Communication (RO-MAN). pp. 873--879. IEEE (2021)

\bibitem{fitzgerald2022inquire}
Fitzgerald, T., Koppol, P., Callaghan, P., Wong, R.Q.J.H., Simmons, R., Kroemer, O., Admoni, H.: Inquire: Interactive querying for user-aware informative reasoning. In: 6th Annual Conference on Robot Learning (2022)

\bibitem{gasteiger2023factors}
Gasteiger, N., Hellou, M., Ahn, H.S.: Factors for personalization and localization to optimize human--robot interaction: A literature review. International Journal of Social Robotics  15(4),  689--701 (2023)

\bibitem{ghafurian2021social}
Ghafurian, M., Hoey, J., Dautenhahn, K.: Social robots for the care of persons with dementia: a systematic review. ACM Transactions on Human-Robot Interaction (THRI)  10(4),  1--31 (2021)

\bibitem{gombolay2018fast}
Gombolay, M.C., Wilcox, R.J., Shah, J.A.: Fast scheduling of robot teams performing tasks with temporospatial constraints. IEEE Transactions on Robotics  34(1),  220--239 (2018)

\bibitem{Habibian2022Heres}
Habibian, S., Jonnavittula, A., Losey, D.P.: Here’s what i’ve learned: Asking questions that reveal reward learning. J. Hum.-Robot Interact.  11(4) (Sep 2022), \url{https://doi.org/10.1145/3526107}

\bibitem{hansen2016cma}
Hansen, N.: The cma evolution strategy: A tutorial. arXiv preprint arXiv:1604.00772  (2016)

\bibitem{hansen2010comparing}
Hansen, N., Auger, A., Ros, R., Finck, S., Po{\v{s}}{\'\i}k, P.: Comparing results of 31 algorithms from the black-box optimization benchmarking bbob-2009. In: Proceedings of the 12th annual conference companion on Genetic and evolutionary computation. pp. 1689--1696 (2010)

\bibitem{hansen2003reducing}
Hansen, N., M{\"u}ller, S.D., Koumoutsakos, P.: Reducing the time complexity of the derandomized evolution strategy with covariance matrix adaptation (cma-es). Evolutionary computation  11(1),  1--18 (2003)

\bibitem{jeon2020shared}
Jeon, H.J., Losey, D.P., Sadigh, D.: Shared autonomy with learned latent actions. arXiv preprint arXiv:2005.03210  (2020)

\bibitem{keselman2023optimizing}
Keselman, L., Shih, K., Hebert, M., Steinfeld, A.: Optimizing algorithms from pairwise user preferences. In: 2023 IEEE/RSJ International Conference on Intelligent Robots and Systems (IROS). pp. 4161--4167. IEEE (2023)

\bibitem{kian2024can}
Kian, M.J., Zong, M., Fischer, K., Singh, A., Velentza, A.M., Sang, P., Upadhyay, S., Gupta, A., Faruki, M.A., Browning, W., et~al.: Can an llm-powered socially assistive robot effectively and safely deliver cognitive behavioral therapy? a study with university students. arXiv preprint arXiv:2402.17937  (2024)

\bibitem{lee2005can}
Lee, K.M., Park, N., Song, H.: Can a robot be perceived as a developing creature? effects of a robot's long-term cognitive developments on its social presence and people's social responses toward it. Human communication research  31(4),  538--563 (2005)

\bibitem{lu2022preference}
Lu, S., Zheng, M., Fontaine, M.C., Nikolaidis, S., Culbertson, H.: Preference-driven texture modeling through interactive generation and search. IEEE transactions on haptics  15(3),  508--520 (2022)

\bibitem{martins2019user}
Martins, G.S., Santos, L., Dias, J.: User-adaptive interaction in social robots: A survey focusing on non-physical interaction. International Journal of Social Robotics  11,  185--205 (2019)

\bibitem{mataric2016socially}
Matari{\'c}, M.J., Scassellati, B.: Socially assistive robotics. Springer handbook of robotics pp. 1973--1994 (2016)

\bibitem{moro2018learning}
Moro, C., Nejat, G., Mihailidis, A.: Learning and personalizing socially assistive robot behaviors to aid with activities of daily living. ACM Transactions on Human-Robot Interaction (THRI)  7(2),  1--25 (2018)

\bibitem{myers2022learning}
Myers, V., Biyik, E., Anari, N., Sadigh, D.: Learning multimodal rewards from rankings. In: Conference on robot learning. pp. 342--352. PMLR (2022)

\bibitem{nemlekar2023transfer}
Nemlekar, H., Dhanaraj, N., Guan, A., Gupta, S.K., Nikolaidis, S.: Transfer learning of human preferences for proactive robot assistance in assembly tasks. In: Proceedings of the 2023 ACM/IEEE International Conference on Human-Robot Interaction. pp. 575--583 (2023)

\bibitem{nikolaidis2015efficient}
Nikolaidis, S., Ramakrishnan, R., Gu, K., Shah, J.: Efficient model learning from joint-action demonstrations for human-robot collaborative tasks. In: Proceedings of the tenth annual ACM/IEEE international conference on human-robot interaction. pp. 189--196 (2015)

\bibitem{o2024design}
O'Connell, A., Banga, A., Ayissi, J., Yaminrafie, N., Ko, E., Le, A., Cislowski, B., Mataric, M.: Design and evaluation of a socially assistive robot schoolwork companion for college students with adhd. In: Proceedings of the 2024 ACM/IEEE International Conference on Human-Robot Interaction. pp. 533--541 (2024)

\bibitem{perovic2023adaptive}
Perovic, G., Iori, F., Mazzeo, A., Controzzi, M., Falotico, E.: Adaptive robot-human handovers with preference learning. IEEE Robotics and Automation Letters  (2023)

\bibitem{rossi2017user}
Rossi, S., Ferland, F., Tapus, A.: User profiling and behavioral adaptation for hri: A survey. Pattern Recognition Letters  99,  3--12 (2017)

\bibitem{sadigh2017active}
Sadigh, D., Dragan, A., Sastry, S., Seshia, S.: Active preference-based learning of reward functions (2017)

\bibitem{saleh2021robot}
Saleh, M.A., Hanapiah, F.A., Hashim, H.: Robot applications for autism: a comprehensive review. Disability and Rehabilitation: Assistive Technology  16(6),  580--602 (2021)

\bibitem{sharma2022correcting}
Sharma, P., Sundaralingam, B., Blukis, V., Paxton, C., Hermans, T., Torralba, A., Andreas, J., Fox, D.: Correcting robot plans with natural language feedback. arXiv preprint arXiv:2204.05186  (2022)

\bibitem{shi2023evaluating}
Shi, Z., Chen, H., Velentza, A.M., Liu, S., Dennler, N., O'Connell, A., Mataric, M.: Evaluating and personalizing user-perceived quality of text-to-speech voices for delivering mindfulness meditation with different physical embodiments. In: Proceedings of the 2023 ACM/IEEE International Conference on Human-Robot Interaction. pp. 516--524 (2023)

\bibitem{shi2024build}
Shi, Z., O'Connell, A., Li, Z., Liu, S., Ayissi, J., Hoffman, G., Soleymani, M., Matari{\'c}, M.J.: Build your own robot friend: An open-source learning module for accessible and engaging ai education. In: Proceedings of the AAAI Conference on Artificial Intelligence. vol.~38, pp. 23137--23145 (2024)

\bibitem{suguitan2019blossom}
Suguitan, M., Hoffman, G.: Blossom: A handcrafted open-source robot. ACM Transactions on Human-Robot Interaction (THRI)  8(1),  1--27 (2019)

\bibitem{tapus2008user}
Tapus, A., {\c{T}}{\u{a}}pu{\c{s}}, C., Matari{\'c}, M.J.: User—robot personality matching and assistive robot behavior adaptation for post-stroke rehabilitation therapy. Intelligent Service Robotics  1,  169--183 (2008)

\bibitem{venkatesh2000theoretical}
Venkatesh, V., Davis, F.D.: A theoretical extension of the technology acceptance model: Four longitudinal field studies. Management science  46(2),  186--204 (2000)

\bibitem{zhang2017human}
Zhang, J., Fiers, P., Witte, K.A., Jackson, R.W., Poggensee, K.L., Atkeson, C.G., Collins, S.H.: Human-in-the-loop optimization of exoskeleton assistance during walking. Science  356(6344),  1280--1284 (2017)

\bibitem{zhou2021designing}
Zhou, E., Shi, Z., Qiao, X., Matari{\'c}, M.J., Bittner, A.K.: Designing a socially assistive robot to support older adults with low vision. In: Social Robotics: 13th International Conference, ICSR 2021, Singapore, Singapore, November 10--13, 2021, Proceedings 13. pp. 443--452. Springer (2021)

\end{thebibliography}

\appendix
\section{Hyperparameter Sensitivity Analysis}
In this paper we explored using CMA-ES and CMA-ES-IG as algorithms for generating queries. We used the default step size parameter of $\sigma=0.5$ which is recommended when there is little information about the structure of the problem space. This ensures that our results are not tuned specifically for our simulation environment, making it more informative of our user study.

\begin{figure}
    \centering
    \includegraphics[width=\linewidth]{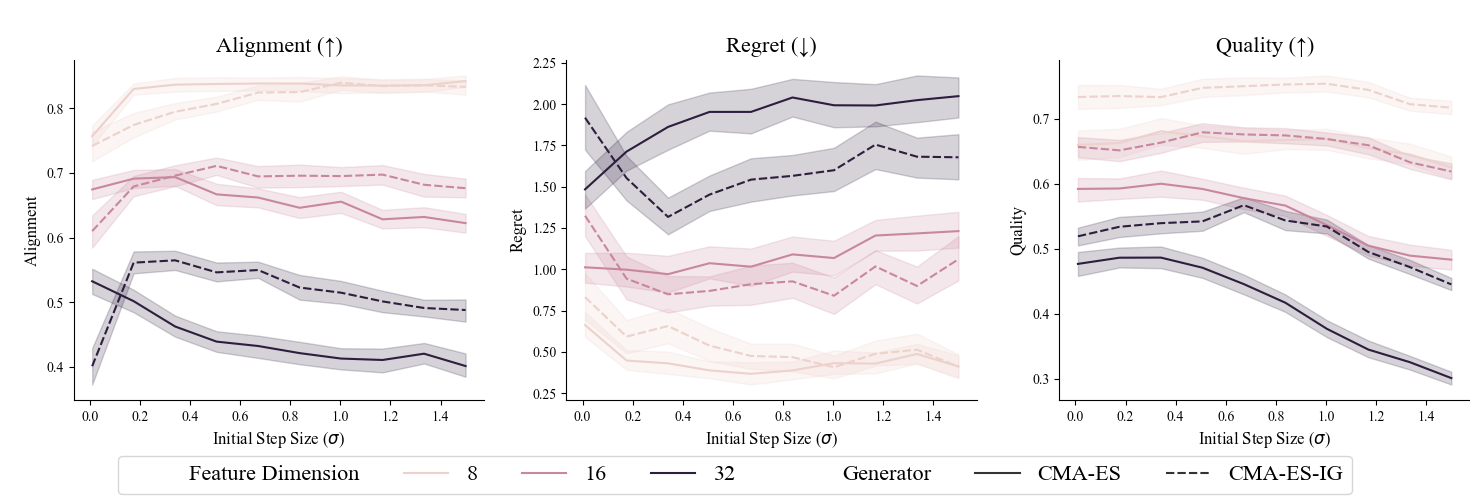}
    \caption{Initial step-size sensitivity. We show the AUC of the labeled metric (corresponding to the values in \autoref{tab:quantitative_results}), across 10 different settings of $\sigma$ ranging from 0.01 to 1.5.  We find that the choice of hyperparameter does not largely influence any of the three metrics of interest.  Across most hyperparameter values, CMA-ES-IG outperforms CMA-ES on Alignment and Regret metrics. Across all hyperparameter values, CMA-ES-IG outperforms CMA-ES on the Quality metric.}
    \label{fig:enter-label}
\end{figure}

We found that CMA-ES is relatively insensitive to changes in parameters due to its automatic step-size adaptation mechanism. This result highlights the strength of CMA-ES-IG to be applicable to a variety of problems without requiring intensive tuning processes.

\end{document}